\documentclass[conference]{IEEEtran}
\IEEEoverridecommandlockouts
\usepackage{cite}
\usepackage{amsmath,amssymb,amsfonts}
\usepackage{algorithmic}
\usepackage{graphicx}
\usepackage{textcomp}
\usepackage{xcolor}
\usepackage{booktabs}
\usepackage{multirow}
\usepackage{subfloat}
\usepackage{subcaption}

\def\BibTeX{{\rm B\kern-.05em{\sc i\kern-.025em b}\kern-.08em
    T\kern-.1667em\lower.7ex\hbox{E}\kern-.125emX}}
\begin{document}

\title{Zero-shot domain adaptation based on dual-level mix and contrast
\thanks{This work is supported by JSPS Grants-in-Aid for Scientific Research and JST CREST}
}

\author{\IEEEauthorblockN{1\textsuperscript{st} Yu Zhe}
\IEEEauthorblockA{\textit{AIP} \\
\textit{RIKEN}\\
Tokyo, Japan \\
zhe.yu@riken.jp}
\and
\IEEEauthorblockN{2\textsuperscript{nd} Jun Sakuma}
\IEEEauthorblockA{\textit{AIP; School of Computing} \\
\textit{RIKEN; Tokyo Institute of Technology }\\
Tokyo, Japan  \\
sakuma@c.titech.ac.jp}}

\maketitle

\begin{abstract}
Zero-shot domain adaptation (ZSDA) is a domain adaptation problem in the situation that labeled samples for a target task (task of interest) are only available from the source domain at training time, but for a task different from the task of interest (irrelevant task), labeled samples are available from both source and target domains. In this situation, classical domain adaptation techniques can only learn domain-invariant features in the irrelevant task. However, due to the difference in sample distribution between the two tasks, domain-invariant features learned in the irrelevant task are biased and not necessarily domain-invariant in the task of interest. To solve this problem, this paper proposes a new ZSDA method to learn domain-invariant features with low task bias. To this end, we propose (1) data augmentation with dual-level mixups in both task and domain to fill the absence of target task-of-interest data, (2) an extension of domain adversarial learning to learn domain-invariant features with less task bias, and (3) a new dual-level contrastive learning method that enhances domain-invariance and less task biasedness of features. Experimental results show that our proposal achieves good performance on several benchmarks.
\end{abstract}

\begin{IEEEkeywords}
Zero-shot domain adaptation, Mix-up, domain adversarial training

\end{IEEEkeywords}

\section{Introduction}
\label{s:intro}
Machine learning models have achieved remarkable performance in various recognition tasks in recent years. In the application of machine learning to real-world problems, one of the common obstacles is domain shift \cite{hendrycks2019benchmarking}, which causes poor generalization performance when there is a gap between the distribution of training data (source domain) and that of test data (target domain). Domain adaptation is an area of research that aims to improve predictive performance in the target domain even in the presence of domain shift and has been studied widely under a variety of situations \cite{ganin2016domain,dual,tang2020discriminative}.

\begin{figure}[t]
 \includegraphics[width=0.8\columnwidth]{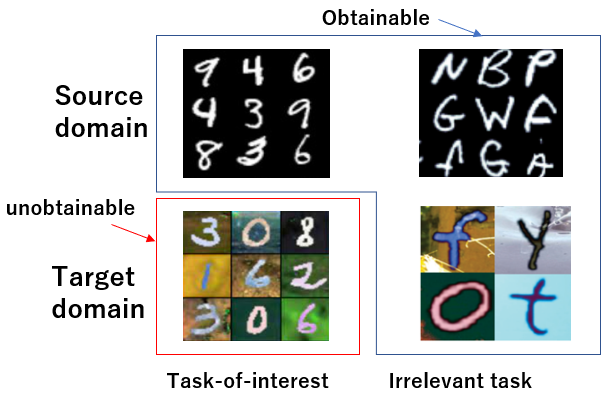}
 \caption{Zero-shot domain adaptation}
\label{digital}
\end{figure}

In this study, we pay attention to Zero-Shot Domain Adaptation (ZSDA) \cite{zsda2018zero}, where (1) we consider two domains: source domain and target domain (e.g., gray-scale images and color images), (2) we consider two different classification tasks, task of interest (ToI, e.g., digit classification) and irrelevant task (IrT, e.g., alphabet classification), (3) the goal is to solve the task of interest in the target domain, but the learner cannot obtain samples of ToI in the target domain for training, (4) instead, the learner can obtain samples of both tasks in the source domain and samples of the irrelevant task in the target domain (Figure \ref{digital}).

ZSDA is useful when (1) domain shifts exist and (2) labeled data on the task of interest are difficult to obtain or costly, but data on an irrelevant task are readily available. For example, consider the following situation in medical image diagnosis. Suppose Hospital A and Hospital B have sets of medical images, but the images of the two hospitals follow different distributions due to the difference in the imaging equipment. Hospital A possesses a set of medical images related to a rare disease (the task of interest), while Hospital B does not. On the other hand, Hospital A and Hospital B both have many medical images related to a common disease (irrelevant task) with many patients. When these samples can be shared with each other, can we obtain a model that can accurately diagnose the suspected rare disease with medical images taken by Hospital B’s imaging equipment? ZSDA provides a solution to problems in such a situation.

\subsection{Related work\label{related work}}
Domain adaptation techniques for addressing domain shifts in various problem settings have been proposed, such as unsupervised domain adaptation \cite{ganin2016domain}, partial domain adaptation \cite{zhang2018importance}, and source-free domain adaptation \cite{kim2021domain}. ZSDA is a special case for them. The difficulty inherent in ZSDA is that, in addition to domain shift, the task we can obtain samples at training time does not match the task we can obtain samples at test time. To solve ZSDA, we need to deal with not only domain shifts but also the gap that arises between different tasks. 

Two main approaches to solving this problem have been known: one is based on generative models, and the other is based on feature disentanglement. The generative model approach uses a generative model such as GAN (Generative Adversarial Network) \cite{gan} with a domain transformation module. Task of interest data in the target domain are synthesized using the generative model from the task of interest data in the source domain through the domain transformation module, and then the training data are augmented with the synthesized samples, which is expected to fill the absence of the ToI data in the target domain. The limitations of this approach are two-fold. First, it requires significant computational resources and careful tuning of hyper-parameters to train a generative model capable of performing domain transformations and producing high-quality images. In addition, it is intrinsically difficult to generate high-quality ToI images without using ToI data in the target domain. If the obtained generative model is of insufficient quality, the predictive performance of the models trained with such synthetic samples would be significantly degraded, too.

Another strategy for solving ZSDA is based on feature disentanglement. Again, there are two main difficulties with ZSDA: domain shift and the gap between the tasks. Domain shift can be solved if one can obtain features invariant to differences in domain with retaining information useful for the classification task \cite{ganin2016domain}. However, even if domain-invariant features could be obtained for one task, they might not necessarily be domain-invariant for another task. In other words, domain invariance of the features needs to be maintained across different tasks to deal with ZSDA. To solve the two difficulties simultaneously, DF-ZSDA \cite{zsda_colla} proposes to force the feature extractor to disentangle domain-related information from task-related information. To this end, DF-ZSDA designed an algorithm with two stages. The first stage utilized dual-level adversarial learning to learn domain-invariant features that are not affected by the difference between the two tasks. As discussed in \cite{zsda_colla}, the features learned at the first stage are still biased by the difference in tasks. To deal with this bias, DF-ZSDA introduces an attention module in the second stage to mitigate the task difference bias left in the features. 

Recently, \cite{fahes2023poda} attempted to utilize large language models (LLMs) to solve zero-shot domain problems. In their setting, information about the target domain is not provided by irrelevant data from the target domain but by a piece of natural language that can be used by LLMs. We will not compare with this approach in our experiments due to differences in settings resulting from whether or not LLMs are used and how information from the target domain is provided.

\subsection{Contribution}
In this paper, we propose an algorithm, termed \textbf{d}ual \textbf{m}ixup \textbf{c}ontrastive \textbf{l}earning (DMCL), to solve the ZSDA problem. DMCL aims to obtain domain-invariant features that can work on ToI. As we already discussed, learning domain-invariant features which are less affected by task differences is necessary to solve the ZSDA problem. With this in mind, the contributions of this paper are summarized as follows.

\begin{enumerate}
    \item We propose to use a dual mixup, which generates samples by randomly interpolating two domains and two tasks. Dual mixup allows to generate various samples belonging to intermediate tasks and domains without the training of generative models.

    \item We propose an extension of domain adversarial training to obtain domain-invariant features that can generalize over ToI by further forcing the model to distinguish the dual mixup samples.
    
    \item Samples generated with dual mixup have intermediate class labels and domain labels that interpolate the two tasks and domains. To exploit the diversity of dual mixup samples to enhance domain invariance and reduce task biasedness of the features, we introduce a novel dual-level contrastive learning method that contrasts pairs of samples at two levels: task and domain.

    \item We experimentally demonstrate that our proposal achieves good performance among several competitors with several datasets. Also, additional experiments verify our proposal is able to learn domain-invariant features for ToI data.
\end{enumerate}

\section{Preliminary}
\subsection{Problem setting}
In the zero-shot domain adaptations (ZSDA) setting, we differentiate data in two ways, domains and tasks. First, let $\mathcal{X}$ be the input space and $\mathcal{Y}$ be the label space. A domain is defined as a joint distribution $P^{XY}$ on $\mathcal{X} \times \mathcal{Y}$  \footnote{We often use $P$ to refer to $P^{XY}$ for simplicity}. In the zero-shot domain adaptation (ZSDA) task, data come from two domains, a source domain $P_{s}$, and a target domain $P_{t}$. Second, in the ZSDA setting, data are also drawn from two tasks, a task of interest (ToI), and an irrelevant task (IrT). ZSDA setting assumes that different classification tasks are distinguished by distinct label sets. That is, ToI and IrT have distinct label sets $\mathcal{C}^{r}$ and $\mathcal{C}^{ir}$, where $\mathcal{C}^{r} \cap \mathcal{C}^{ir} = \emptyset$.

For the ToI, we only have sample-label pairs from the source domain, namely we have 
$D_{s}^{r}=\left\{\left(x_{s_j}^{r}, y_{s_j}^{r}\right)\right\}_{j=1}^{N}$ where $y_s^r \in \mathcal{C}^r$. For the IRT, we have sample-label pairs from both source and target domains, namely we have$D_{s}^{ir}=\left\{\left(x_{s_j}^{ir}, y_{s_j}^{ir}\right)\right\}_{j=1}^{N}$ where $y_s^{ir} \in \mathcal{C}^{ir}$ and $D_{t}^{ir}=\left\{\left(x_{t_j}^{ir}, y_{t_j}^{ir}\right)\right\}_{j=1}^{N}$ where $y_t^{ir} \in \mathcal{C}^{ir}$, respectively. We also use a domain label to indicate the domain information of a sample. For $x_{s}^{r}$ and $x_{s}^{ir}$, the corresponding domain label $d_{s}$ is 0; for $x_{t}^{ir}$ the corresponding domain label $d_{t}$ is 1. Given $D_{s}^{ir}, D_{t}^{ir}, D_{s}^{r}$, the goal of the ZSDA task is to learn a model that can generalize to the distribution of target ToI data. We denote samples drawn from the distribution of target ToI data by $D_{t}^{r}=\left\{\left(x_{t_j}^{r}, y_{t_j}^{r}\right)\right\}_{j=1}^{N}$.

\subsection{Domain adversarial training}
\label{DANN pre}
Domain-invariant features refer to features that are invariant in both source and target domains. Learning domain-invariant features is a classical way to alleviate domain shifts. In this work, we extend Domain adversarial training (DAT) \cite{ganin2016domain} to learn domain-invariant features. Hence, we first introduce DAT.

 Let $\mathcal{X}$ and $\mathcal{Y}$ be the input space and label. $D_s$ is a dataset from the source domain, and $D_t$ is a dataset from the target domain. Let $G: \mathcal{X} \mapsto \mathbb{R}^m$ be the feature extractor, and $C: \mathbb{R}^m \mapsto \mathcal{Y}$ be the category classifiers. Domain adversarial training introduces an additional domain classifier $D: \mathbb{R}^m \mapsto[0,1]$ for distinguishing domain information of input samples where samples from the source domain with domain label 0, and samples from the target domain with domain label 1. The domain adversarial training can be formulated as follows:

$$\min _{G,C} \max _D \mathcal{L}_c(G, C)+\lambda \mathcal{L}_{d}(G, D)$$

where $\mathcal{L}_c(G, C)=\mathbb{E}_{\left(\mathbf{x}_s, y_s\right) \sim D_s} \ell\left(C\left(G\left(\mathbf{x}_s\right)\right), y_s\right),$

$$\begin{aligned}
\mathcal{L}_{d}(G, D)= \mathbb{E}_{\mathbf{x}_s \sim D_s} \log \left(1-D\left(G\left(\mathbf{x}_s\right)\right)\right)+ \\\mathbb{E}_{\mathbf{x}_t \sim D_t} \log D\left(G\left(\mathbf{x}_t\right)\right) .
\end{aligned}$$
Here, $\ell$ is the cross-entropy loss and $\lambda$ is a trade-off hyper-parameter. $G$ and $C$ are both trained to predict category label correctly, hence $G$ is expected to provide discriminative features. $L_{d}$ is used to train a domain classifier. $D$ is trained to predict domain labels of input samples correctly, while $G$ is trained to provide features that could make $D$ misclassify. With this procedure, the features obtained with $G$ are expected to be domain-invariant. $L_{c}$ is a classification loss for the label classifier.

\subsection{Mixup}
\label{s:mixup}
Mixup is a data augmentation technique that generates virtual sample-label pairs by a convex combination of two sample-label pairs. In this paper, we utilize mixup to generate virtual samples to act as target ToI data. Category-level mixup \cite{mixup} performs data augmentation by randomly interpolating a pair of samples with distinct labels. It can be formulated as follows:

$$\begin{aligned}
\widetilde{\mathbf{x}} &=\mathcal{M}_\lambda\left(\mathbf{x}_i, \mathbf{x}_j\right)=\lambda \mathbf{x}_i+(1-\lambda) \mathbf{x}_j ,\\
\widetilde{y} &=\mathcal{M}_\lambda\left(y_i, y_j\right)=\lambda y_i+(1-\lambda) y_j\\
\end{aligned}$$where $\lambda \sim \operatorname{Beta}(\alpha, \alpha)$ for $\alpha \in(0, \infty)$. Recently, both empirical \cite{mixup} and theoretical \cite{whymixup} results show that category-level mixup can improve the model's generalization ability.

Inspired by \cite{mixup}, \cite{xu2020adversarial} proposed the domain-level mixup to improve the discriminative ability of the domain classifier. Domain-level mixup applies a convex combination on samples from different domains and their domain label. 
$$\widetilde{\mathbf{x}} =\mathcal{M}_\lambda\left(\mathbf{x}_i, \mathbf{x}_j\right)=\lambda \mathbf{x}_i+(1-\lambda) \mathbf{x}_j,$$
$$\widetilde{d} =\mathcal{M}_\lambda\left(d_i, d_j\right)=\lambda d_i+(1-\lambda) d_j$$where $\lambda \sim \operatorname{Beta}(\alpha, \alpha)$, for $\alpha \in(0, \infty)$.

\section{Methodology}
\subsection{Overview}
\begin{figure*}[t]
\centering
\includegraphics[width=1.8\columnwidth]{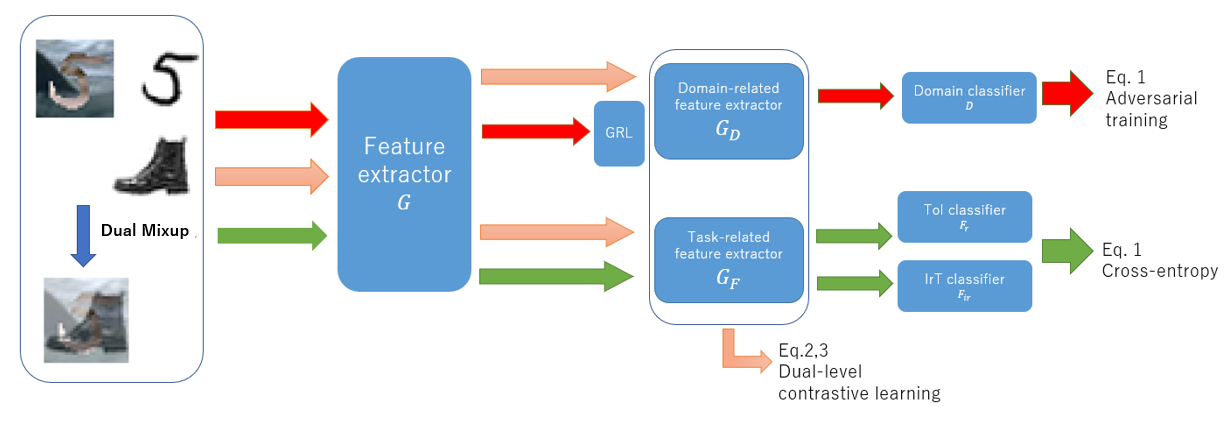}  
\caption{An illustration of our proposal DMCL. Here GRL refers to a gradient reversal layer. The red arrow indicates the adversarial learning procedure, the green arrow indicates a standard training procedure, and the pink arrow indicates the contrastive learning objectives.} 
\label{fig:overview}
\end{figure*}

In this section, we propose a learning algorithm, termed \textbf{d}ual \textbf{m}ixup \textbf{c}ontrastive \textbf{l}earning (DMCL), for the zero-shot domain adaptation. As discussed in section \ref{s:intro}, to solve the ZSDA task, DMCL aims to learn domain-invariant features that generalize over ToI task. An overview of our proposal is shown in Figure \ref{fig:overview}.

First, we extend mixup to synthesize intermediate samples between domains and tasks to fill the absence of target ToI data. Then, we extend domain adversarial training with intermediate samples to learn domain-invariant features. As shown in Figure \ref{fig:overview}, our model structure is an extension of domain adversarial neural network with a domain classifier $D\circ G_{D}$, and two task classifiers, $F_r\circ G_{F}$ for ToI, $F_{ir} \circ G_{F}$ for IrT. With adversarial learning (red arrow in Figure \ref{fig:overview}), the feature extractor $G$ is expected to extract domain-invariant features with maintaining discriminative capability in both ToI and IrT. Here, the domain invariance of the features is induced by the gradient reversal layer (GRL) which reverses the gradient by multiplying a negative value during the backpropagation. Also, the discriminative capability of the features is due to training by cross-entropy loss (green arrow in Figure \ref{fig:overview}).

As we demonstrate with ablation studies later, domain-invariant features learned by domain adversarial training are insufficient to generalize to ToI. Since we can only have IrT data from both domains, the resulting features are biased toward the IrT task. To deal with this task bias, we design dual-level contrastive learning objectives (pink arrow in Figure \ref{fig:overview}). With the first contrastive learning objective, $G_{F}$ is encouraged to be more sensitive to differences in tasks and class labels while caring less about differences in domains. This further enhances the domain invariance of the features input to $F_{r}$ and $F_{ir}$. Similarly, with the second contrastive learning objective, $G_{D}$ focuses on domain-related information more while ignoring task-related information. Hence, the performance of the domain classifier will be further improved.

Moreover, since $G_{D}$ only learns domain-related features, when $G$ is adversarially trained with gradient information from $G_{D}$, only domain-related features are eliminated, and the task-related features which are helpful in classification tasks are unaffected. On the other hand, with adversarial training, $G$ tends to learn domain-invariant features, i.e., the input of $G_{F}$ becomes domain-invariant. This leads to a better domain invariance of outputs of $G_{F}$. Therefore, by training features with the contrastive learning objective and the domain adversarial training objective alternately, domain-invariance and task-unbiasedness can enhance each other.

\subsection{Dual Mixup for intermediate samples}
In this work, we provide intermediate data from different domains and tasks to act as target ToI data with low computation costs. This strategy has third merits. First, these intermediate data contain information from both source and target domains, also from both ToI and IrT. A model trained on these data should be able to learn features that work on different domains and tasks. Second, we can generate intermediate data without using target ToI data since intermediate data between source ToI data and target IrT already contain information from two domains and tasks. Third, we synthesize intermediate samples by mixup. Mixup only requires applying a convex combination on different samples; hence this method requires low computation resources.

To synthesize intermediate data between both tasks and domains, we extend the mixup \cite{mixup} technique. Formally, for $x_{i}$ in $D_{s}^{r}$ and $x_{j}$ in $D^{ir}=D_{s}^{ir} \cup D_{t}^{ir}$, we synthesize virtual data by $\widetilde{x} =\mathcal{M}_\lambda\left(x_i, x_j\right)$.

Then, unlike single-level mixup, we mix both the category label and the domain label for $\widetilde{x}$ by:
$$\widetilde{y} =\mathcal{M}_\lambda\left(y_i, y_j\right), \quad \widetilde{d} =\mathcal{M}_\lambda\left(d_i, d_j\right)$$
where $\lambda \sim \operatorname{Beta}(\alpha, \alpha)$, for $\alpha \in(0, \infty)$.

\subsection{Domain adversarial training with dual mixup}
Using data augmentation with intermediate data synthesized by dual mixup, we extend the domain adversarial training method for learning domain-invariant features. During this procedure, we force the domain classifier and label classifier to distinguish samples generated by mixing samples in different domains and tasks with random proportions. Through this training, the feature extractor is trained to be able to handle data from different domains and tasks.

Specifically, we define $C_{d}= D \circ G_{D} \circ G$, which means $C_{d}(x)=D(G_{D}(G(x)))$. Similarly, we define $C_{r}= F_r \circ G_{F} \circ G$ and $C_{ir}= F_{ir} \circ G_{F} \circ G$. Then, the domain adversarial training with dual mixup samples is shown in the following: 

\begin{multline}  
\min _{\substack{G, G_{F} \\ F_r, F_{ir}}} \max _{G_{D},D} \mathcal{L}_{adv}= \mathcal{L}_{d}(C_{d}) + \mathcal{L}_{md}(C_{d}) + \\ \mathcal{L}_f(C_{r}, C_{ir}) +  \mathcal{L}_{mf}(C_{r},C_{ir}) .
\label{adversarial loss}    
\end{multline}

where
\begin{equation*}
\resizebox{0.9\hsize}{!}{$\mathcal{L}_{d}(C_{d})=\mathbb{E}_{\mathbf{x}_s \sim D_{s}} \log \left(1- C_{d}(\mathbf{x}_s)\right)+\mathbb{E}_{\mathbf{x}_t \sim D_{t}^{ir}} \log C_{d}(\mathbf{x}_t),$}
\end{equation*}

\begin{equation*} 
\resizebox{0.9\hsize}{!}{$\mathcal{L}_{md}(C_{d})=\mathbb{E}_{\substack{\mathbf{x}_{i} \sim D_{s} \\ \mathbf{x}_{j} \sim D_{t}^{ir} \\ \lambda \sim \operatorname{Beta}(\alpha, \alpha)}}  \lambda \log (1-C_{d}(\widetilde{x}))+ (1-\lambda) \log C_{d}(\widetilde{x})
.$}
\end{equation*}  
\begin{multline*}
\label{ce_loss}
\resizebox{0.9\hsize}{!}{$\mathcal{L}_{f}(C_{r},C_{ir})=\mathbb{E}_{\left(\mathbf{x}, y\right) \sim D_s^r} \ell (C_{r}(\mathbf{x}), 
y) + \mathbb{E}_{\left(\mathbf{x}, y\right) \sim D^{ir}} \ell (C_{ir}(\mathbf{x}), y)$}
\end{multline*}

\begin{multline*} 
\mathcal{L}_{mf}(C_{r},C_{ir})=\mathbb{E}_{\substack{ \lambda \sim \operatorname{Beta}(\alpha, \alpha)\\ \mathbf{x}_{i} \sim D_s^r, \mathbf{x}_{j} \sim D^{ir}  }} \lambda   \ell (C_{r}(\widetilde{x}), y_i)
+ \\(1-\lambda)  \ell (C_{ir}(\widetilde{x}), y_{ir}).
\end{multline*} Here, $D_{s}=D_{s}^{ir} \cup D_{s}^r$, $D^{ir}=D_{s}^{ir} \cup D_{t}^{ir}$, $\widetilde{x} =\mathcal{M}_\lambda\left(x_i, x_j\right)$, $\ell$ is the cross-entropy loss, and $\alpha$ is hyper-parameter.

In the above training objective, $\mathcal{L}_{d}$ and $\mathcal{L}_{md}$ force $G$ to learn domain-invariant features and force $D$ and $G_{D}$ to give high domain classification accuracy. $\mathcal{L}_f$ and $\mathcal{L}_{mf}$ are the classification error of ToI, IrT data, and of their mixup, respectively; this forces $C_{r}$ and $C_{ir}$ become to be able to classify ToI and IrT data, respectively.

\subsection{Dual contrastive learning for disentanglement}
\label{sec:contra}
We then design two contrastive learning objectives to enhance domain invariance and reduce task biasedness in the features. We assume that the features extracted from an image can be divided into two types: domain-related features and task-related features. We expect the domain-related features to contain information that allows identifying the domain while it is insensitive to changes in the task and category. In contrast, we expect the task-related features to contain information that allows identifying category labels while it is insensitive to changes in the domain. As discussed in section \ref{s:intro}, disentangling task-related features from domain-related features helps the model in classifying target ToI data.

We introduce two feature extractors, $G_{F}$ and $G_{D}$, to realize the feature disentanglement. $G_{F}$ is used to extract task-related features, $G_{D}$ is used to extract domain-related features. To enforce feature disentanglement, our key intuition is that when two samples from different domains but with the same category are fed into $G_{F}$, their corresponding outputs should be very similar since these two samples are only different at the domain level. Likewise, when two samples from different categories but with the same domain are fed into $G_{D}$, their corresponding outputs should be the same.  

\begin{figure}[t]
\centering
 \includegraphics[width=0.75\columnwidth]{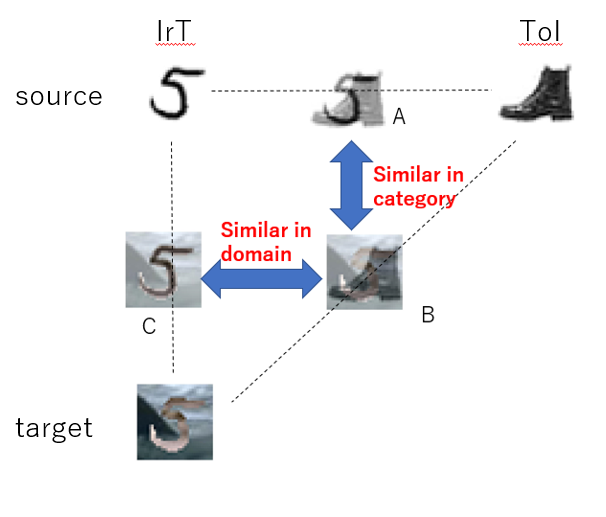}
 \caption{An illustration of dual contrastive learning, here the dashed line indicates apply mixup on two samples.}
\label{dual}
\end{figure}

Considering that we have no way of knowing how domain-related and task-related features are mixed together in target ToI data, we need our model to be able to have the generalized feature disentangled capability. This means our model should be able to achieve feature disentanglement for data containing different mixes of domain-related and task-related features. On the other hand, by varying $\lambda$, mixup is able to generate intermediate data containing different proportions of domain information and task information. Hence, we utilize intermediate data to realize our intuition for obtaining generalized feature disentangled ability. Figure \ref{dual} explains a high-level concept of the proposed mixup procedure.

Formally, let us consider three mini-batches of $K$ samples $X_s^r, X_s^{ir}, X_t^{ir}$ from $D_s^r$, $D_s^{ir}$, $D_t^{ir}$, We first apply mixup on any pair of min-batches of samples from the three mini-batches with the same $\lambda$ for obtaining intermediate samples.

$$\mathbf{A} = \left\{\mathcal{M}_\lambda\left(\mathbf{X}_{s}^{r_i}, \mathbf{X}_{s}^{{ir}_i}\right)\right\}_{i=1}^{K}, \mathbf{B}= \left\{\mathcal{M}_\lambda\left(\mathbf{X}_{s}^{r_i}, \mathbf{X}_{t}^{{ir}_i}\right)\right\}_{i=1}^{K}$$

$$\mathbf{C} = \left\{\mathcal{M}_\lambda\left(\mathbf{X}_{s}^{{ir}_i}, \mathbf{X}_{t}^{{ir}_i}\right)\right\}_{i=1}^{K}$$

Then, we get three mini-batches of mixup samples $\mathbf{A},\mathbf{B},\mathbf{C}$. We assume that after applying mixup, the corresponding category information and the domain information will also be mixed in equal proportions. Then, $\mathbf{A}_{i}$ and $\mathbf{B}_{i}$ contain the same category information but different domain information, and $\mathbf{B}_{i}$ and $\mathbf{C}_{i}$ contain the same domain information but different category information. 

Considering that the core idea of our intuition is similar to the goal of contrastive learning, our intuition can be naturally realized by contrastive learning. We take $G_{D}$ as an example to explain our detailed approach. For $G_{D}$, our key idea suggests that $G_{D}(G(\mathbf{B}_{i}))$ and $G_{D}(G(\mathbf{C}_{i}))$ should become as similar as possible, which means $(\mathbf{B}_{i},\mathbf{C}_{i})$ should be treated as a positive pair in contrastive learning; we treat the other $2(K-1)$ augmented samples within a minibatch and $\mathbf{B}_{i}$ as $2(K-1)$ negative pairs, like did in \cite{simclr}. 

Then, given positive and negative pairs, we utilize the normalized temperature-scaled cross entropy  (NT-Xent) loss $\mathcal{L}_{con_d}$ to train $G_{D}$:

\begin{equation}
\label{con_d}
 \mathcal{L}_{con_d}\left( G_{D}\right)=-\log \frac{\operatorname{exp} \left(\operatorname{sim}\left(\mathbf{zd}_\mathbf{B}^i, \mathbf{zd}_\mathbf{C}^i\right)/ \tau\right)}{\sum_{\substack{j=1, j \neq i \\ T \in\{\mathbf{B}, \mathbf{C}\}}}^K \operatorname{exp} \left(\operatorname{sim}\left(\mathbf{zd}_\mathbf{B}^i, \mathbf{zd}_T^j\right)/ \tau\right)}   
\end{equation} where, $\operatorname{sim}(\boldsymbol{u}, \boldsymbol{v})=\boldsymbol{u}^{\top} \boldsymbol{v} /\|\boldsymbol{u}\|\|\boldsymbol{v}\|$ denote the dot product between $l_{2}$ normalized $\boldsymbol{u}$ and $\boldsymbol{v}$. $\mathbf{zd}_\mathbf{B}^{i}=G_{D}(G(\mathbf{B}_{i}))$, $\mathbf{zd}_\mathbf{C}^{i}=G_{D}(G(\mathbf{C}_{i}))$, $\tau$ is the temperature parameter.

Likewise, $G_{F}$ treats $(\mathbf{A}_{i},\mathbf{B}_{i})$ as positive pair. $G_{F}$ is trained by NT-Xent loss $L_{con_f}$: 
\begin{equation}
\label{con_f}
\mathcal{L}_{con_f}\left(G_{F}\right)=-\log \frac{\operatorname{exp} \left(\operatorname{sim}\left(\mathbf{zf}_\mathbf{A}^i, \mathbf{zf}_\mathbf{B}^i\right)/ \tau\right)}{\sum_{\substack{j=1, j \neq i \\ T \in\{\mathbf{A}, \mathbf{B}\}}}^K \operatorname{exp} \left(\operatorname{sim}\left(\mathbf{zf}_\mathbf{A}^i, \mathbf{zf}_T^j\right)/ \tau\right)}
\end{equation}
where, $\mathbf{zf}_\mathbf{A}^{i}=G_{F}(G(\mathbf{A}_{i}))$, $\mathbf{zf}_\mathbf{B}^{i}=G_{F}(G(\mathbf{B}_{i}))$.

The entire training procedure is performed as follows. We first optimize equation \eqref{adversarial loss}, then optimize equation \eqref{con_d} and \eqref{con_f} with shared intermediate samples. We repeat the above two steps alternately.

\section{Experiments}
\subsection{Experimental results on two benchmarks}
\textbf{Datasets} We evaluate our proposal on two benchmarks. The first is X-NIST, which consists of four domains and four classification tasks. X-NIST is based on four datasets, including MNIST (task M) \cite{mnist}, Fashion-MNIST (task F) \cite{fmnist}, EMNIST (task E) \cite{emnist}, NIST (task N) \cite{nist}. Images in these datasets are all in the gray-scale domain (domain G). To test the domain adaptation performance, we create the color (domain C), edge (domain E), and negative domains (domain N). The color domain is synthesized by using \cite{ganin2016domain}'s method, blending the samples with randomly selected patches from the BSDS500 dataset \cite{bsds500}. The edge domain is created by applying the canny edge detector, and the negative domain is obtained by subtracting the original pixel value from 255. When conducting experiments, we choose two of the four tasks as ToI and IrT, but we do not consider the NIST and EMNIST combinations since their label spaces are not completely different.

The second benchmark is Office-Home datasets \cite{officehome}. It consists of images from four different domains: Artistic images (Ar), Clip images (Cl), Product images (Pr), and Real-world images (Rw). This dataset contains images from 65 object categories for each domain. When conducting experiments, we used 10 random categories from 65 categories as ToI and the rest as the IrT.
 
\textbf{Implementation deatails} In all experiments, the classifier $F_r$, $F_{ir}$, and $D$ were implemented with one fully connected layer. In the X-NIST benchmark, two feature extractors, $G_{D}$ and $G_{F}$, were implemented with three convolutional layers. $G$ was implemented with three convolutional layers. We set the batch size as 64 and the total number of iterations as 7,000. In the Office-Home dataset, we utilized ResNet-50 \cite{resnet} pre-trained on ImageNet. $G$ was implemented with stages 0 to 3 of ResNet-50. Stage 4 of ResNet-50 was copied into two parts: $G_{D}$ and $G_{F}$. We set the batch size as 32 and the total number of iterations as 15,000. For all tasks, we used Adam with a learning rate of 0.0002. In half of the full training iterations, the learning rate was decayed by 0.1. More details are shown in the supplementary.

\textbf{Comparison methods} We compare our proposal with two types of works: (1) the generative model-based methods, ZDDA \cite{zsda2018zero}, CoCoGAN \cite{zsda2019conditional}, and Wang2020 \cite{zsda2020adversarial}. (2) Feature disentanglement-based method, DF-ZSDA \cite{zsda_colla} \footnote{We used the official code provided by the author of DF-ZSDA to reproduce DF-ZSDA's experimental results on X-NIST. The reproduced results are worse than the results reported in their paper. The experimental results on office-home datasets are drawn from their paper because the provided code is not prepared for office-home.}. For all competitors, the model was trained on source ToI, source IrT, and target IrT. Then the model was tested on the target ToI data.

\begin{table*}[t]
\caption{Experimental result on X-NIST with classification accuracy (\%) averaged over 10 runs}
\centering
\tabcolsep=0.11cm
\scalebox{1.0}{\begin{tabular}{@{}l|ll|lll|lll|ll|ll|l@{}}
\toprule
\multirow{2}{*}{Domains}           & \multicolumn{1}{l|}{\multirow{2}{*}{Methods}} & ToI & \multicolumn{3}{l|}{MNIST($D_M$)}              & \multicolumn{3}{l|}{FashionMNIST($D_{F}$)}     & \multicolumn{2}{l|}{NIST($D_N$)} & \multicolumn{2}{l|}{EMNIST($D_E$)} & \multirow{2}{*}{Average} \\ \cmidrule(lr){3-13}
                                   & \multicolumn{1}{l|}{}                         & IrT & $D_F$         & $D_N$         & $D_E$         & $D_M$         & $D_N$         & $D_E$         & $D_M$          & $D_F$          & $D_M$            & $D_F$           &                          \\ \midrule
\multirow{5}{*}{G$\rightarrow$ C} & \multicolumn{2}{l|}{ZDDA}                           & 73.2          & 92.0          & 94.8          & 51.6          & 43.9          & 65.3          & 34.3           & 21.9           & 71.2             & 47.0            & 59.5                     \\
                                   & \multicolumn{2}{l|}{CoCoGAN}                        & 78.1          & 92.4          & \textbf{95.6} & 56.8          & 56.7          & \textbf{66.8} & 41.0           & 44.9           & 75.0             & 54.8            & 66.2                     \\
                                   & \multicolumn{2}{l|}{Wang2020}                       & \textbf{81.2} & \textbf{93.3} & 95.0          & \textbf{57.4} & 58.7          & 62.0          & 44.6           & 45.5           & 72.4             & 58.9            & \textbf{66.9}            \\
                                   & \multicolumn{2}{l|}{DF-ZSDA }                        & 68.7          & 77.0          & 86.3          & 40.2          & 42.3          & 42.4          & 45.7           & 31.3           & \textbf{83.0}    & \textbf{66.8}   & 58.4                     \\
                                   & \multicolumn{2}{l|}{Ours}                           & 76.6          & 89.2          & 80.9          & 45.3          & \textbf{59.5} & 57.0          & \textbf{50.2}  & \textbf{59.2}  & 78.1             & 60.9            & 65.7                     \\ \midrule
\multirow{5}{*}{G$\rightarrow$ E} & \multicolumn{2}{l|}{ZDDA}                           & 72.5          & 91.5          & 93.2          & 54.1          & 54.0          & 65.8          & 42.3           & 28.4           & 73.6             & 50.7            & 62.6                     \\
                                   & \multicolumn{2}{l|}{CoCoGAN}                        & 79.6          & 94.9          & 95.4          & 61.5          & 57.5          & 71.0          & 48.0           & 36.3           & 77.9             & 58.6            & 68.1                     \\
                                   & \multicolumn{2}{l|}{Wang2020}                       & 81.4          & 93.5          & \textbf{96.3} & \textbf{63.2} & 58.7          & \textbf{72.4} & 49.9           & 38.6           & 78.2             & 61.1            & 69.3                     \\
                                   & \multicolumn{2}{l|}{DF-ZSDA}                        & 79.5          & \textbf{95.5} & 93.5          & 33.4          & 30.7          & 35.8          & \textbf{53.4}  & \textbf{47.0}  & \textbf{85.5}    & \textbf{74.4}   & 62.9                     \\
                                   & \multicolumn{2}{l|}{Ours}                           & \textbf{87.0} & 91.5          & 93.1          & 59.9          & \textbf{63.8} & 64.7          & 48.3           & 44.2           & 78.5             & 71.5            & \textbf{70.3}            \\ \midrule
\multirow{5}{*}{G$\rightarrow$ N} & \multicolumn{2}{l|}{ZDDA}                           & 77.9          & 82.4          & 90.5          & 61.4          & 47.4          & 62.7          & 37.8           & 38.7           & 76.2             & 53.4            & 62.8                     \\
                                   & \multicolumn{2}{l|}{CoCoGAN}                        & 80.3          & 87.5          & 93.1          & 66.0          & 52.2          & 69.3          & 45.7           & 53.8           & 81.1             & 56.5            & 68.6                     \\
                                   & \multicolumn{2}{l|}{Wang2020}                       & -             & -             & -             & -             & -             & -             & -              & -              & -                & -               & -                        \\
                                   & \multicolumn{2}{l|}{DF-ZSDA}                        & 59.7          & 81.0  & 90.6          & 68.7          & 64.3          & 77.6          & 58.7           & 59.0             & 77.7             & 64.0            & 70.1                    \\
                                   & \multicolumn{2}{l|}{Ours}                           & \textbf{94.6} & \textbf{94.2}          & \textbf{97.6} & \textbf{69.8} & \textbf{68.7} & \textbf{78.9} & \textbf{62.7}  & \textbf{64.9}  & \textbf{86.2}    & \textbf{86.4}   & \textbf{80.4}            \\ \midrule
\multirow{5}{*}{C $\rightarrow$ G} & \multicolumn{2}{l|}{ZDDA}                           & 67.4          & 85.7          & 87.6          & 55.1          & 49.2          & 59.5          & 39.6           & 23.7           & 75.5             & 52.0            & 59.5                    \\
                                   & \multicolumn{2}{l|}{CoCoGAN}                        & 73.2          & 89.6          & 94.7          & 61.1          & 50.7          & 70.2          & 47.5           & 57.7           & 80.2             & 67.4            & 69.2                    \\
                                   & \multicolumn{2}{l|}{Wang2020}                       & 73.7          & 91.0          & 93.4          & 62.4          & 53.5          & 71.5          & 50.6           & 58.1           & 83.5             & 70.9            & 70.9                    \\
                                   & \multicolumn{2}{l|}{DF-ZSDA}                        & \textbf{98.1} & \textbf{99.1} & \textbf{99.1} & \textbf{88.0} & \textbf{89.1} & \textbf{89.5} & \textbf{69.0}    & \textbf{69.1}  & \textbf{91.3}    & \textbf{92.1}   & \textbf{88.4}           \\
                                   & \multicolumn{2}{l|}{Ours}                           & 92.1          & 90.3          & 92.8          & 86.2          & 76.2          & 74.9          & 65.9           & 62.8           & 89.4             & 75.9            & 80.7                    \\ \midrule
\multirow{5}{*}{N$\rightarrow$ G} & \multicolumn{2}{l|}{ZDDA}                           & 78.5          & 90.7          & 87.6          & 56.6          & 57.1          & 67.1          & 34.1           & 39.5           & 67.7             & 45.5            & 62.4                     \\
                                   & \multicolumn{2}{l|}{CoCoGAN}                        & 80.1          & 92.8          & 93.6          & 63.4          & 61.0          & 72.8          & 47.0           & 43.9           & 78.8             & 58.4            & 69.2                     \\
                                   & \multicolumn{2}{l|}{Wang2020}                       & 82.6          & \textbf{94.6}          & 95.8          & 67.0          & 68.2          & 77.9          & 51.1           & 44.2           & 79.7             & 62.2            & 72.3                     \\
                                   & \multicolumn{2}{l|}{DF-ZSDA}                        & 64.1          & 68.7 & 89.5          & 58.7          & 57.2   & 30.3          & 58.4           & 51.0            & 73.4             & 56.6            & 60.8                     \\
                                  & \multicolumn{2}{l|}{Ours}& \textbf{95.8} & 92.4& \textbf{97.9}& \textbf{75.0} & \textbf{73.9}     & \textbf{78.1} & \textbf{64.6}  & \textbf{57.2}  & \textbf{88.3}    & \textbf{87.5}   & \textbf{81.1}            \\ \bottomrule
\end{tabular}}
\label{bench_results}
\end{table*}

\begin{table*}[t]
\caption{Experimental result on Office-Home with classification accuracy (\%) averaged over 10 runs}
\tabcolsep=0.11cm
\centering
\scalebox{1.0}{\begin{tabular}{@{}l|lll|lll|lll|lll@{}}
\toprule
Source   & \multicolumn{3}{l|}{Pr}                       & \multicolumn{3}{l|}{Rw}                       & \multicolumn{3}{l|}{Ar}                       & \multicolumn{3}{l}{Cl}                        \\ \midrule
Target   & Ar            & Cl            & Rw            & Ar            & Cl            & Pr            & Cl            & Pr            & Rw            & Ar            & Pr            & Rw            \\
CoCoGAN  & 57.6          & 53.4          & 71.7          & 69.2          & 51.3          & 65.8          & 62.3          & 69.5          & 74.5          & 66.7          & 74.0          & 66.4          \\
Wang2020 & \textbf{70.3} & 60.8          & 74.8          & 72.2          & 61.4          & 72.2          & 62.7          & 71.9          & 76.3          & \textbf{72.6} & \textbf{75.1} & 73.9          \\
DF-ZSDA  & 64.4          & \textbf{69.2} & \textbf{82.0} & \textbf{77.9} & \textbf{76.2} & \textbf{88.5} & 71.0          & 76.5          & \textbf{85.1} & 62.1          & 68.7          & \textbf{75.1} \\
Ours     & 67.5          & 65.1          & 78.9          & 74.3          & 69.0          & 75.8          & \textbf{72.1} & \textbf{76.7} & 83.8          & 69.8          & 73.0          & 71.5          \\ \bottomrule
\end{tabular}}
\label{office_1}
\end{table*}

\textbf{Results}
Table \ref{bench_results} shows the prediction accuracy of the target ToI data for over 10 task combinations on the X-NIST. Overall, these results show that our proposal exhibits a good domain adaptation ability in the ZSDA setting; our proposal achieves the best average accuracy on X-NIST. In particular, our proposal achieves the best performance when domain shifts happen between the gray-scale domain and the negative domain (the third and fifth columns in Table \ref{bench_results}). This is because our proposal relies on the mixup technique, which assumes that intermediate domains can be obtained by linear interpolating source and target domains. Gray-scale and negative domains satisfy this assumption very well, so our proposal performs particularly well. When domain shifts are $C\rightarrow G$, our proposal achieves the second-best result, only worse than DF-ZSDA. When domain shifts are $G\rightarrow C$ and $ G\rightarrow E$ (the first and second columns in Table \ref{bench_results}), the performance gap between our proposal and the best method, Wang2020, is within four percent. This gap is acceptable when considering that our proposal does not require heavy computation resources to train a generative model. 

Table \ref{office_1} shows the prediction accuracy of the target ToI data on the office-home. Although the domain shift in this benchmark is more complex than X-NIST, our proposal still shows good performance to overcome domain shift. Compared with generative model-based methods, our proposal achieves competitive or better results with fewer training resources. Compared with DF-ZSDA, our method achieves close results in certain situations. However, when the source domain is Rw and the target domains are Cl or Pr (the sixth and seventh columns in Table \ref{office_1}), there is a clear gap between our approach and DF-ZSDA. We suspect that the reason for not performing well is that the gap between domains is complex. This complexity weakens the effect of the mixup and therefore affects the performance of our proposal.

\subsection{Visualizations}
\begin{figure}
     \centering
     \begin{subfigure}[b]{0.4\columnwidth}
         \centering
         \includegraphics[width=\columnwidth]{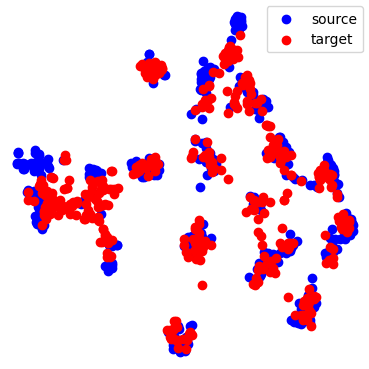}
         \caption{Illustration of domain-invariance}
     \end{subfigure}
     \quad
     \begin{subfigure}[b]{0.4\columnwidth}
         \centering
         \includegraphics[width=\columnwidth]{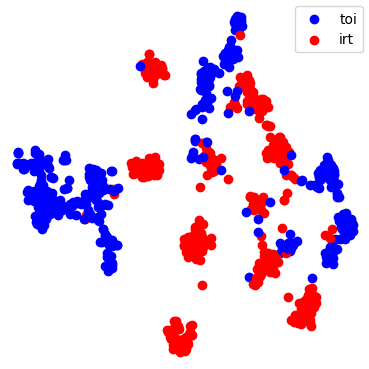}
         \caption{Illustration of contrasting task}
     \end{subfigure}\\     
     \begin{subfigure}[b]{0.4\columnwidth}
         \centering
         \includegraphics[width=\columnwidth]{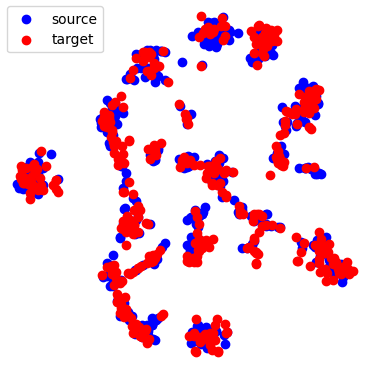}
         \caption{Illustration of domain-invariance}
     \end{subfigure}
     \quad
     \begin{subfigure}[b]{0.4\columnwidth}
         \centering
         \includegraphics[width=\columnwidth]{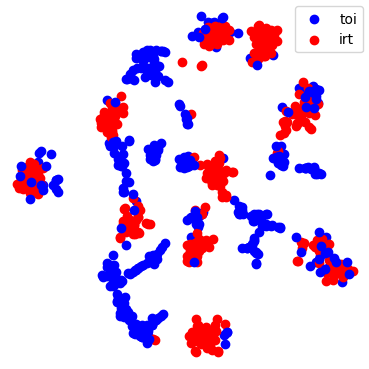}
         \caption{Illustration of contrasting task}
     \end{subfigure}
\caption{Visualization results when domain shift is domain G to domain N. At the top, the ToI is MNIST, IrT is Fashion-MNIST. At the bottom, the ToI is MNIST, IrT is EMNIST.}
\label{visualize_2}
\end{figure}

In this work, we are ultimately concerned with the features used to classify ToI data, which are outputs of $G_{F}$. We expect $G_{F}$ can learn domain-invariant features and is able to separate features belonging to different tasks. To verify whether our proposal works as expected, we visualize the feature space of $G_{F}$ on X-NIST. More specifically, we randomly select 200 samples from source IrT, source ToI, target IrT, and target ToI datasets, respectively. Then, we utilize t-SNE \cite{tsne} to visualize features extracted by $G_{F}$ on a two-dimensional space. In Figure \ref{visualize_2}, we use two ways to colorize feature points. First, we colorize points by their domain label. The feature distributions of source and target domains are very similar. This indicates $G_{F}$ can extract domain-invariant features. Then, we colorized points by their task. We can find that the features of ToI and IrT can be separated, which indicates our contrastive learning objective works well. With the above observation, we concluded that our proposal works as expected. Visualizations on more datasets are shown in the supplementary due to space limits.

\subsection{Ablation studies}
To check the importance of each module in DMCL, we perform the following ablation studies on the X-NIST.
\begin{table}[t]
\centering
\caption{Ablation studies on X-NIST with classification accuracy (\%) averaged over 10 runs}
\scalebox{1.0}{\begin{tabular}{l|lllll}
\hline
Domain shift             & G$\rightarrow$ C & G$\rightarrow$ E & G$\rightarrow$ N & C$\rightarrow$ G & N$\rightarrow$ G \\ \hline
w/o dual mixup           & 59.9             & 63.1             & 65.2             & 74.2             & 76.0             \\
w/o dual contrastive     & 56.4             & 66.2             & 48.4             & 75.0             & 61.5             \\
Ours                     & \textbf{65.7}    & \textbf{70.3}    & \textbf{80.4}     & \textbf{80.7}    & \textbf{81.1}    \\ \hline
\end{tabular}}
\label{t:ab}
\end{table}

\textbf{Importance of the dual mixup} We extend domain adversarial training with mixup samples. To verify this module is necessary, we remove the mixup samples used in domain adversarial training. The second row in Table \ref{t:ab} shows the averaged classification accuracy over 10 task combinations of the target ToI data when we remove dual mixup module. Overall, these results show that the dual mixup module is important. When this module is removed, the difference in performance can be up to 17\%.

\textbf{Importance of the dual-level contrastive learning} We utilize dual-level contrastive learning to force feature disentanglement between domain-related features and task-related features. To verify this module is necessary, we remove two contrastive learning objectives. The third row in Table \ref{t:ab} shows that dual-level contrastive learning is important, and only domain adversarial training is not enough to solve ZSDA. For all domain shifts, removing contrastive learning objectives leads to worse accuracy on the target domain. This fits our expectation. Exploring the intrinsic relationships with contrastive objectives is helpful in solving ZSDA tasks. More ablation studies are shown in the supplementary.

\section{Conclusion}
In this paper, we propose DMCL to learn domain-invariant features that are not affected by the difference between tasks. Specifically, we design a dual mixup to synthesize intermediate samples between tasks and domains. These data bridge the gap of target ToI data. Then we design two contrastive learning objectives to explore the relationship among data from two aspects: domain and task. With contrastive learning, our model is able to separate domain-related features and task-related features. Finally, by applying adversarial learning, the resulting domain-invariant features are not affected by differences in tasks. In the evaluation, DMCL shows good performance in solving zero-shot domain adaptation problems. Moreover, we visualize our proposal on feature space and confirm our proposal works as expected. 

\section*{Acknowledgment}
This work is partly supported by JSPS KAKENHI Grant number JP23H00483 and JST CREST Grant number JPMJCR21D3.

\bibliographystyle{unsrt}
\bibliography{egbib}

\begin{thebibliography}{10}

\bibitem{hendrycks2019benchmarking}
Dan Hendrycks and Thomas Dietterich.
\newblock Benchmarking neural network robustness to common corruptions and perturbations.
\newblock {\em arXiv preprint arXiv:1903.12261}, 2019.

\bibitem{ganin2016domain}
Yaroslav Ganin, Evgeniya Ustinova, Hana Ajakan, Pascal Germain, Hugo Larochelle, Fran{\c{c}}ois Laviolette, Mario Marchand, and Victor Lempitsky.
\newblock Domain-adversarial training of neural networks.
\newblock {\em The journal of machine learning research}, 17(1):2096--2030, 2016.

\bibitem{dual}
Yuan Wu, Diana Inkpen, and Ahmed El-Roby.
\newblock Dual mixup regularized learning for adversarial domain adaptation.
\newblock In {\em European Conference on Computer Vision}, pages 540--555. Springer, 2020.

\bibitem{tang2020discriminative}
Hui Tang and Kui Jia.
\newblock Discriminative adversarial domain adaptation.
\newblock In {\em Proceedings of the AAAI Conference on Artificial Intelligence}, volume~34, pages 5940--5947, 2020.

\bibitem{zsda2018zero}
Kuan-Chuan Peng, Ziyan Wu, and Jan Ernst.
\newblock Zero-shot deep domain adaptation.
\newblock In {\em Proceedings of the European Conference on Computer Vision (ECCV)}, pages 764--781, 2018.

\bibitem{zhang2018importance}
Jing Zhang, Zewei Ding, Wanqing Li, and Philip Ogunbona.
\newblock Importance weighted adversarial nets for partial domain adaptation.
\newblock In {\em Proceedings of the IEEE conference on computer vision and pattern recognition}, pages 8156--8164, 2018.

\bibitem{kim2021domain}
Youngeun Kim, Donghyeon Cho, Kyeongtak Han, Priyadarshini Panda, and Sungeun Hong.
\newblock Domain adaptation without source data.
\newblock {\em IEEE Transactions on Artificial Intelligence}, 2(6):508--518, 2021.

\bibitem{gan}
Ian Goodfellow, Jean Pouget-Abadie, Mehdi Mirza, Bing Xu, David Warde-Farley, Sherjil Ozair, Aaron Courville, and Yoshua Bengio.
\newblock Generative adversarial networks.
\newblock {\em Communications of the ACM}, 63(11):139--144, 2020.

\bibitem{zsda_colla}
Won~Young Jhoo and Jae-Pil Heo.
\newblock Collaborative learning with disentangled features for zero-shot domain adaptation.
\newblock In {\em Proceedings of the IEEE/CVF International Conference on Computer Vision}, pages 8896--8905, 2021.

\bibitem{fahes2023poda}
Mohammad Fahes, Tuan-Hung Vu, Andrei Bursuc, Patrick P{\'e}rez, and Raoul de~Charette.
\newblock Poda: Prompt-driven zero-shot domain adaptation.
\newblock In {\em Proceedings of the IEEE/CVF International Conference on Computer Vision}, pages 18623--18633, 2023.

\bibitem{mixup}
Hongyi Zhang, Moustapha Cisse, Yann~N. Dauphin, and David Lopez-Paz.
\newblock mixup: Beyond empirical risk minimization.
\newblock In {\em International Conference on Learning Representations}, 2018.

\bibitem{whymixup}
Linjun Zhang, Zhun Deng, Kenji Kawaguchi, Amirata Ghorbani, and James Zou.
\newblock How does mixup help with robustness and generalization?
\newblock In {\em International Conference on Learning Representations}, 2021.

\bibitem{xu2020adversarial}
Minghao Xu, Jian Zhang, Bingbing Ni, Teng Li, Chengjie Wang, Qi~Tian, and Wenjun Zhang.
\newblock Adversarial domain adaptation with domain mixup.
\newblock In {\em Proceedings of the AAAI Conference on Artificial Intelligence}, volume~34, pages 6502--6509, 2020.

\bibitem{simclr}
Ting Chen, Simon Kornblith, Mohammad Norouzi, and Geoffrey~E. Hinton.
\newblock A simple framework for contrastive learning of visual representations.
\newblock {\em CoRR}, abs/2002.05709, 2020.

\bibitem{mnist}
Li~Deng.
\newblock The mnist database of handwritten digit images for machine learning research.
\newblock {\em IEEE Signal Processing Magazine}, 29(6):141--142, 2012.

\bibitem{fmnist}
Han Xiao, Kashif Rasul, and Roland Vollgraf.
\newblock Fashion-mnist: a novel image dataset for benchmarking machine learning algorithms.
\newblock {\em CoRR}, abs/1708.07747, 2017.

\bibitem{emnist}
Gregory Cohen, Saeed Afshar, Jonathan Tapson, and Andr{\'{e}} van Schaik.
\newblock {EMNIST:} an extension of {MNIST} to handwritten letters.
\newblock {\em CoRR}, abs/1702.05373, 2017.

\bibitem{nist}
Patrick~J Grother.
\newblock Nist special database 19.
\newblock {\em Handprinted forms and characters database, National Institute of Standards and Technology}, 10, 1995.

\bibitem{bsds500}
Pablo Arbelaez, Michael Maire, Charless Fowlkes, and Jitendra Malik.
\newblock Contour detection and hierarchical image segmentation.
\newblock {\em IEEE transactions on pattern analysis and machine intelligence}, 33(5):898--916, 2010.

\bibitem{officehome}
Hemanth Venkateswara, Jose Eusebio, Shayok Chakraborty, and Sethuraman Panchanathan.
\newblock Deep hashing network for unsupervised domain adaptation.
\newblock In {\em Proceedings of the IEEE conference on computer vision and pattern recognition}, pages 5018--5027, 2017.

\bibitem{resnet}
Kaiming He, Xiangyu Zhang, Shaoqing Ren, and Jian Sun.
\newblock Deep residual learning for image recognition.
\newblock In {\em Proceedings of the IEEE conference on computer vision and pattern recognition}, pages 770--778, 2016.

\bibitem{zsda2019conditional}
Jinghua Wang and Jianmin Jiang.
\newblock Conditional coupled generative adversarial networks for zero-shot domain adaptation.
\newblock In {\em Proceedings of the IEEE/CVF International Conference on Computer Vision}, pages 3375--3384, 2019.

\bibitem{zsda2020adversarial}
Jinghua Wang and Jianmin Jiang.
\newblock Adversarial learning for zero-shot domain adaptation.
\newblock In {\em European Conference on Computer Vision}, pages 329--344. Springer, 2020.

\bibitem{tsne}
Laurens Van~der Maaten and Geoffrey Hinton.
\newblock Visualizing data using t-sne.
\newblock {\em Journal of machine learning research}, 9(11), 2008.

\end{thebibliography}

\vspace{12pt}
\color{red}

\end{document}